\documentclass[11pt]{article}
\usepackage[final]{nlp4call}
\usepackage[utf8]{inputenc}
\usepackage[T1]{fontenc}
\usepackage{latexsym}
\usepackage{subfigure}
\usepackage{graphicx}
\usepackage{times}
\usepackage{inconsolata}
\usepackage{listings}

\definecolor{delim}{RGB}{20,105,176}
\definecolor{numb}{RGB}{106, 109, 32}
\definecolor{string}{rgb}{0.64,0.08,0.08}

\lstdefinelanguage{json}{
    frame=single,
    rulecolor=\color{black},
    showspaces=false,
    showtabs=false,
    inputencoding=latin1,
    extendedchars=true,
    breaklines=true,
    postbreak=\raisebox{0ex}[0ex][0ex]{\ensuremath{\color{gray}\hookrightarrow\space}},
    breakatwhitespace=true,
    basicstyle=\ttfamily\small,
    upquote=true,
    morestring=[b]",
    stringstyle=\color{string},
    literate=
     *{0}{{{\color{numb}0}}}{1}
      {ü}{{\"u}}{1} {ç}{{\c{c}}}{1} {ä}{{\"a}}{1}  {İ}{{\"I}}{1} {ê}{{\^e}}{1}
      {1}{{{\color{numb}1}}}{1}
      {2}{{{\color{numb}2}}}{1}
      {3}{{{\color{numb}3}}}{1}
      {4}{{{\color{numb}4}}}{1}
      {5}{{{\color{numb}5}}}{1}
      {6}{{{\color{numb}6}}}{1}
      {7}{{{\color{numb}7}}}{1}
      {8}{{{\color{numb}8}}}{1}
      {9}{{{\color{numb}9}}}{1}
      {\{}{{{\color{delim}{\{}}}}{1}
      {\}}{{{\color{delim}{\}}}}}{1}
      {[}{{{\color{delim}{[}}}}{1}
      {]}{{{\color{delim}{]}}}}{1},
}

\title{OmniLingo: Listening- and speaking-based language learning}

\author{Francis M. Tyers \\
  Department of Linguistics \\
  Indiana University \\
  Bloomington, IN \\
  {\tt ftyers@iu.edu} \\\And
  Nicholas Howell \\
  Unaffiliated \\
  {\tt nlhowell@gmail.com} \\
  }

\date{}

\begin{document}
\maketitle
\begin{abstract}
In this demo paper we present OmniLingo, an architecture for distributing
data for listening- and speaking-based language learning applications and 
a demonstration client built using the architecture. The architecture is 
based on the Interplanetary Filesystem (IPFS) and puts at the forefront 
user sovereignty over data.
\end{abstract}


\section{Introduction}

Language learning apps can give a fun, convenient way to learn a new language. Like many uses of the internet, though, 
there is an opportunity for dark patterns to sneak in: user data hoarding, targeted presentation, and majority-cultural 
filter bubbles can have negative social impacts, and properietary or closed-source software and always-connected 
centralised backends can create unstable infrastructure and restrict user freedom.

In this paper we present OmniLingo --- an free/open-source project to build language learning protocols, software, and 
infrastructure that avoids these problems, prioritising language communities and user sovereignty over their own data. OmniLingo
is designed to be scaleable, both in terms of infrastructure --- it should not require large server farms to run, and in 
terms of language coverage --- it should not require a large amount of additional effort per language added.

The remainder of the paper is structured as follows: Section~\ref{sec:architecture} gives an overview of the architecture
behind OmniLingo; Section~\ref{sec:data} describes the language data that is used; Section~\ref{sec:tasks} describes
a demonstration user experience given the platform; 
and Section~\ref{sec:future} highlights some future directions for development.


\section{Architecture}\label{sec:architecture}

OmniLingo language data is stored on IPFS\footnote{InterPlanetary File System \cite{benet:14}} in a 
hierarchy of JSON\footnote{JavaScript Object Notation} and MP3 files. The \textit{root index} of a language data store is a JSON
dictionary mapping ISO-639 language codes to \textit{language indices}
and \textit{language metadata} (see Figure~\ref{fig:rootindex}).

\subsection{IPFS}

The InterPlanetary File System (IPFS) is a peer-to-peer
content-addressed filesystem protocol and network~\cite{benet:14}. File
content stored on IPFS is identified by a \textit{content-ID} (CID), a
hash generated by an extensible hashing algorithm designed for
content-addressable networking. Since the address is generated from
the content, modifications to files result in new CIDs.

IPFS network members can generate the CID for a file and then
advertise it; other peers will be able to download directly from the
hoster or indirectly through other peers on the network.

There are two major implementations of IPFS available, the canonical
implementation in Go and a newer partial implementation in JavaScript.

\subsection{OmniLingo data structures}

\begin{figure}
\begin{lstlisting}[language=json]
{
  "or": {
    "cids": [
      "Qm" #...
    ],
    "meta": "Qm" #...
  },
  "pa-IN": {
    "cids": [
      "Qm" #...
    ],
    "meta": "Qm" #...
  }
}
\end{lstlisting}
\caption{An OmniLingo root index. The root index is a JSON dictionary
mapping ISO-639 language codes to language entries. Each language
entry contains a list of references to language indices (see
Figure~\ref{fig:langindex}) and a reference to a language metadata
structure (Figure~\ref{fig:langmeta}). References are IPFS content-ID
multihashes. IPFS CIDs start with \texttt{Qm}, and have been elided
for space.}
\label{fig:rootindex}
\end{figure}

\begin{figure}
\begin{lstlisting}[language=json]
{
  "alternatives": {
    "İ": [
      "I"
    ]
  },
  "display": "Türkçe"
}
\end{lstlisting}
\caption{An OmniLingo language metadata structure. Currently language
metadata consists of display names and reverse-keymaps.}
\label{fig:langmeta}
\end{figure}

The language metadata structure (Figure~\ref{fig:langmeta}) consists
of a ``display name'' for the language and a set of character rewrite
rules to make typing easier. Language indices
(Figure~\ref{fig:langindex}) are JSON lists of clip structures
consisting of MP3 audio sample and difficulty metadata, used to generate
an appropriate exercise for the learner's level. Each clip structure
contains a reference to the sentence (including licence metadata, see
Figure~\ref{fig:sent}), as well as to additional sentence metadata to
inform exercise generation (Figure~\ref{fig:sentmeta}).

\begin{figure}
\begin{lstlisting}[language=json]
[
  {
    "chars_sec": 15.2116,
    "clip_cid": "Qm", #...
    "length": 6.048,
    "sentence_cid": "Qm" #...
    "meta_cid": "Qm", #...
  },
  {
    "chars_sec": 15.97,
    "clip_cid": "Qm", #...
    "length": 5.76,
    "sentence_cid": "Qm" #...
    "meta_cid": "Qm", #...
  },
]
\end{lstlisting}
\caption{An OmniLingo language index is a JSON list of clip
structures. Each clip structure contains some basic metadata about the
clip (duration, characters-per-second) and references to the clip MP3,
a sentence structure (Figure~\ref{fig:sent}), and a sentence
metadata structure (Figure~\ref{fig:sentmeta}). References are IPFS
content-ID multihashes, and have been elided for space.}
\label{fig:langindex}
\end{figure}

\begin{figure}
\begin{lstlisting}[language=json]
{
  "content": "Tavaliselt ongi nii, et
mesinik jääb oma surnud mesilastega
ja mitte mingit lahendust ei tule.",
  "copyright": "CC0-1.0",
  "language": "et"
}
\end{lstlisting}
\caption{An OmniLingo sentence structure, consisting of a
JSON dictionary of original sentence transcript, licence data, and
ISO-639 language code.}
\label{fig:sent}
\end{figure}

\begin{figure}
\begin{lstlisting}[language=json]
{
  "sentence_cid": "Qm", #...
  "tags": [
  	# ...
  ],
  "tokens": [
  	# ...
  ]
}
\end{lstlisting}
\caption{An OmniLingo sentence metadata structure, consisting of a
reference to the sentence structure (see Figure~\ref{fig:sent}),
tokenisation of the sentence and matched token tags. Token tags are
currently either \texttt{"X"} (no tag) or \texttt{"PUNCT"}. The
reference is an IPFS content-ID multihash, and has been elided for
space, as have the tokens and tags.}
\label{fig:sentmeta}
\end{figure}

Root indexes are encouraged to be published to IPNS, so that clients
can receive updates.

\section{Language data}\label{sec:data}

Language data for the project comes primarily from Common Voice \cite{ardila:19}, 
a project run by the Mozilla Foundation that collects voice data for training
speech recognition systems. For a given language, interested parties provide 
sentences and participants in the project read them out. This data is then recorded
and released to the public under a Creative Commons CC-0 licence every three months. The data
is provided in a tab-separated format for transcripts and metadata and MP3 files for
the audio data, as such implementing the format for a language not in Common Voice is a simple
matter of following the same format.

As of writing there are over 100 languages represented in Common Voice, and over 24,210
hours of recordings. The dataset downloads include both the audio files, in MP3
format and the transcripts for each file, along with a certain amount of demographic
information such as gender, age range and accent or variant.

By default OmniLingo extracts 10,000 clips for each language, grouping them into
10 buckets by a given difficulty metric. The current metric used is characters per second, that
is number of characters in the transcript divided by the number of seconds of audio. The
motivation behind this metric is that slower speakers should be easier to understand. However
working on improved difficulty metrics is an ongoing area of research (see §\ref{sec:future} for 
discussion).

Transcripts are processed by a separate library \emph{commonvoice-utils},\footnote{\url{https://github.com/ftyers/commonvoice-utils}} 
which provides tokenisation for the languages in Common Voice and rudimentary
tagging of word tokens (those which can be substituted by a gap) versus punctuation.

\section{Tasks and progression}\label{sec:tasks}

\begin{figure}
\begin{subfigure}[An example question]{
\includegraphics[width=.45\textwidth]{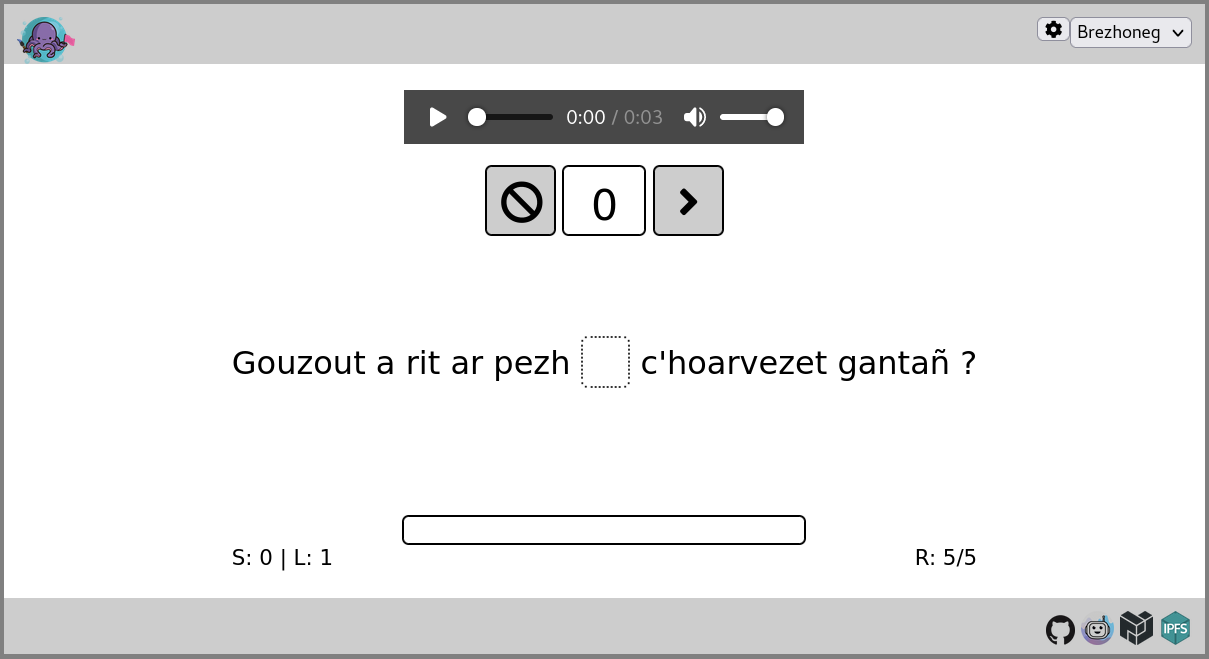}
\label{fig:interface:ex1}
}
\end{subfigure}

\begin{subfigure}[Feedback after a wrong answer]{
\includegraphics[width=.45\textwidth]{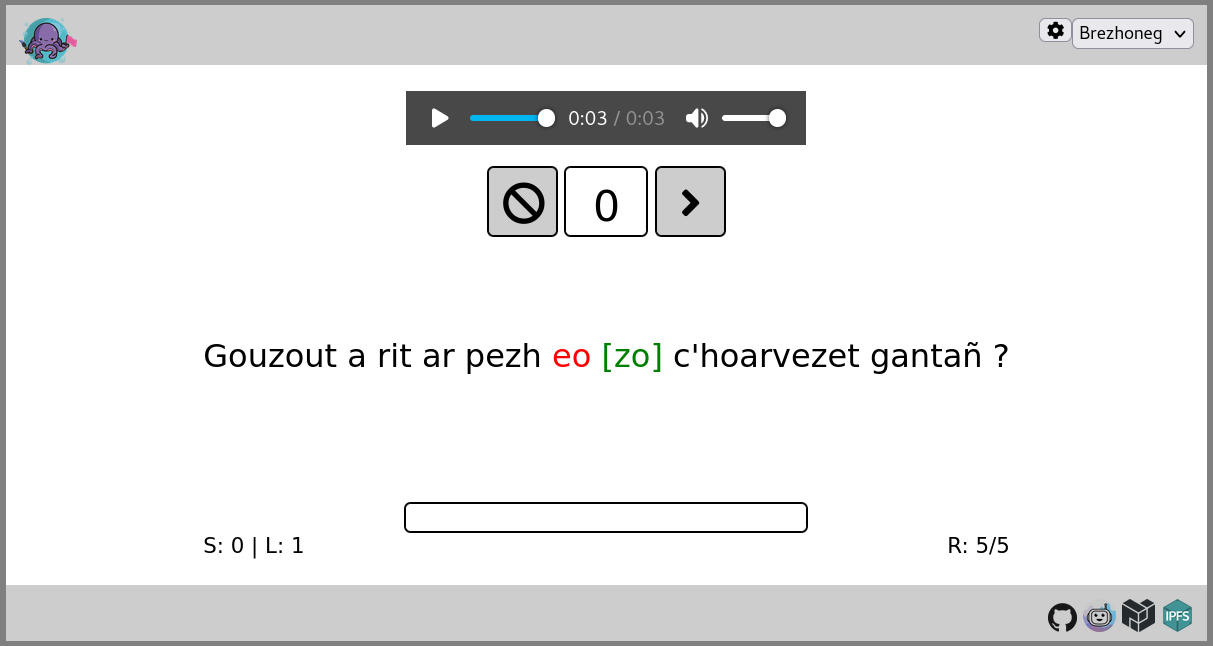}
\label{fig:interface:ex2}
}
\end{subfigure}

\begin{subfigure}[A correct answer]{
\includegraphics[width=.45\textwidth]{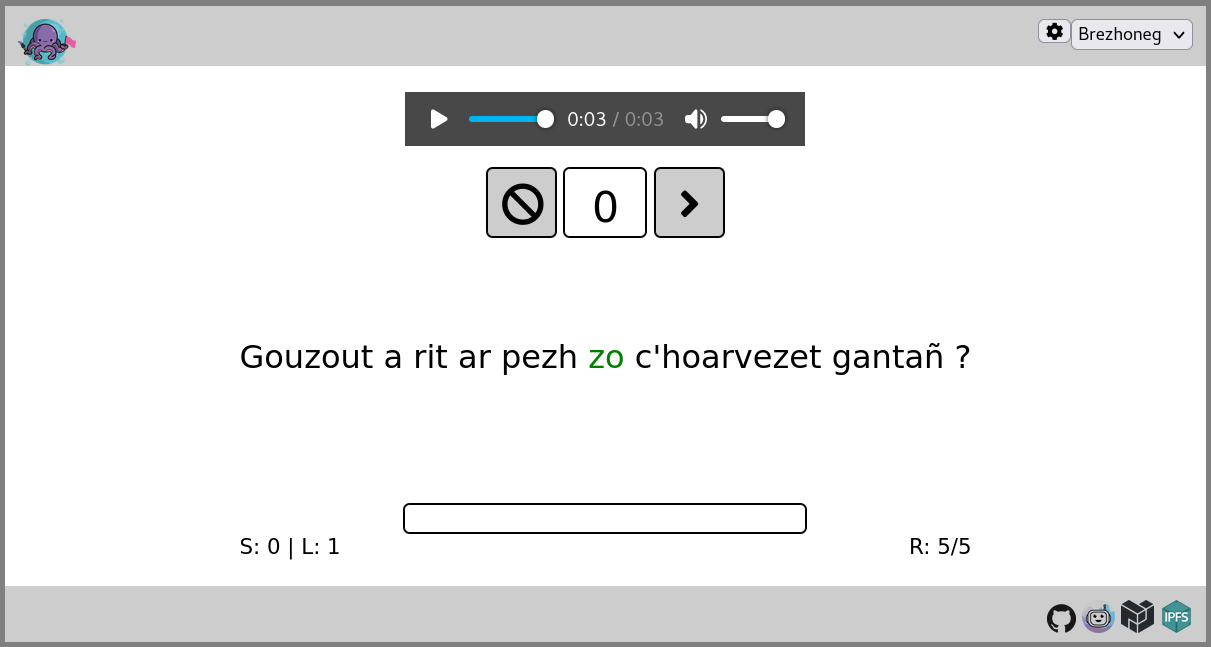}
\label{fig:interface:ex3}
}
\end{subfigure}
\caption{A demonstration interface for OmniLingo showing the gap-filling 
  task in Breton. The sentence \emph{Gouzout a rit ar pezh zo c'hoarvezet gantañ?} `Do you know 
  who arrived with him?' has a gap for the word \emph{zo} `is'. In Figure~\ref{fig:interface:ex2} the learner has made a 
  mistake and written \emph{eo} (another form of `is') and this is corrected to \emph{zo}.} \label{fig:interface}
\end{figure}

Currently the main task type in OmniLingo's demonstration interface is a gap filling task. A learner is presented
with a sentence (see Figure~\ref{fig:interface:ex1}) where one of the words is 
replaced by a gap. The learner is invited to listen to the audio by clicking on the play
button and then fill in the gap according to what she heard. Once the audio has 
finished playing a timer starts to keep track of how long it takes her to fill in the gap.
She may re-listen to the audio file as many times as she wishes.

If the learner fills in a gap incorrectly, she is given the correct answer (Figure~\ref{fig:interface:ex2}) 
and may listen to the clip again or move on to the next clip. If she answers correctly (Figure~\ref{fig:interface:ex3}),
the correct answer is highlighted and again she may move onto the next clip.

Tasks are presented in groups of five, which constitute a level. A learner \emph{passes} a level
when she answers all five tasks in less time than the clip takes to play. Her score is the 
how long it took her to answer subtracted from the total length of the audio at that level, meaning that
faster response times result in higher scores.

At any point the learner may discard (or deactivate) a clip by clicking on the discard button, 
a new random clip from the same bucket is then added to the current group. The user may also choose
to skip a particular clip by clicking on the [>] button. 

Each time a clip is presented the gap is generated randomly from the word tokens. This means that if 
a user listens to a clip and cannot fill in the gap, the next time the sentence is presented the 
gap will likely be in a different place.

The interface indicates which level the learner is currently in \emph{L:} to the bottom left, 
what her current score is \emph{S:} to the bottom left and how many clips are remaining at this 
level \emph{R:} to the bottom right. 

\section{Pronunciation feedback}
Modern speech recognition architectures allow for the design of
pronunciation assistants, analysing input speech and comparing it
per-phone against the speech it was trained on~\cite{papareo}. We have
implemented a pronunciation assistant
based on models trained from the speech data we used in our OmniLingo
indices.

The reference implementation is based on the Coqui STT platform,\footnote{\url{https://github.com/coqui-ai/STT/}}
but it could be adapted to use any browser-based speech recognition system,
for example \emph{Whisper}.\footnote{\url{https://github.com/pluja/whishper}}
\begin{table}
\begin{tabular}{ll}
Tr: & foi classificada para a mostra de talentos \\
Hyp: & foi clacificada para mosta letitãntos \\
Alig:  & foi cla··ificada par··a most·a ·e·t···ntos \\
\end{tabular}
\caption{Example of alignment (Alig) between the transcript (Tr) and the 
ASR hypothesis (Hyp), where a gap is indicated with the middot, ·. 
The sentence is in Portuguese and means ``She qualified for the talent show.''}\label{table:alig}
\end{table}

The speech recognition models are stored on IPFS and are indexed through a list of \texttt{models} 
included in the language metadata structure.

\begin{figure}
\begin{lstlisting}[language=json]
{
 "format":"coqui",
 "licence":"AGPL-3.0",
 "model":"Qm", # ...
 "src":"https://example.com/models/",
 "type":"acoustic"
}
\end{lstlisting}
\caption{An example of the metadata structure for speech recognition models. In 
  this case the format is Coqui STT, the type is acoustic model (for speech recognition)
  and the model is found at CID ``model".}\label{lst:modelinfo}
\end{figure}

\begin{figure}
\begin{lstlisting}[language=json]
{
 "alternatives":{},
 "display":"Português",
 "models":["Qm"] # ...
}
\end{lstlisting}
\caption{A list of language models is included in the language metadata structure. For reference
  see Figure~\ref{fig:langmeta}.}\label{lst:lang2}
\end{figure}

When OmniLingo is loaded, the speech recognition model (in this case 45M) is downloaded and 
stored in localStorage. The speech recorded by the user is stored in the browser 
and transcoded using the WebAudio API.

The system works as follows: The learner is presented with a sentence, they are asked
to record themselves saying the sentence, which they can do any number of times using
the record button. When they have finished, they press the get feedback button.

The recording they have made is run through the speech recognition system. The output
of this system is a hypothesis transcript. This hypothesis is then aligned using the 
Needleman-Wunch algorithm \cite{needleman-wunsch}. The alignment produced contains 
gaps where the hypothesis does not match the transcript. 

\begin{figure}
\centering
\begin{subfigure}[Recording interface]{
\includegraphics[width=0.45\textwidth]{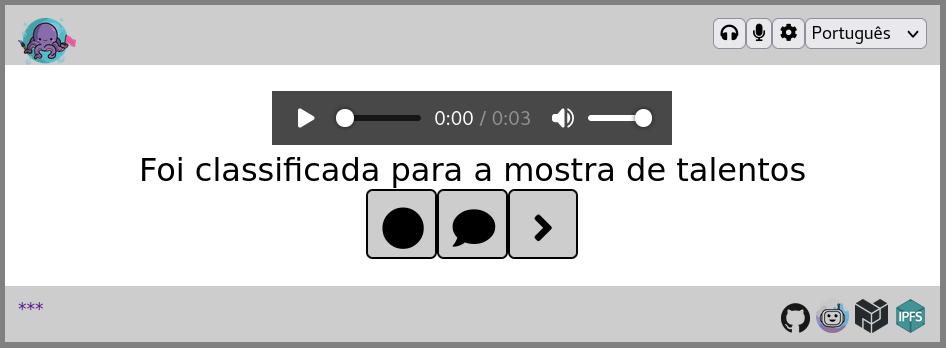}
}
\end{subfigure}
\begin{subfigure}[Requesting feedback]{
\includegraphics[width=0.45\textwidth]{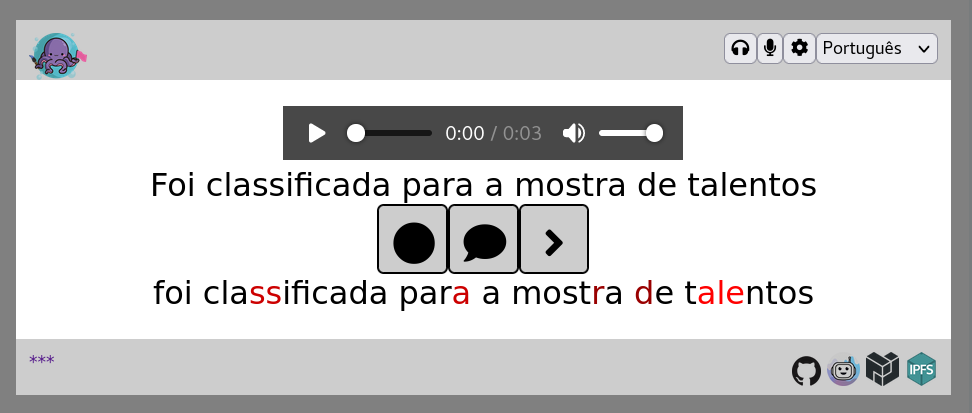}
}
\end{subfigure}
\caption{Example of the pronunciation feedback interface, with original sentence (above) and 
feedback shown (below). The buttons are record, get feedback and skip. }\label{fig:pronunciation}
\end{figure}

In the interface, gaps are filled with the characters from the corresponding position
in the transcript and coloured red to indicate that the learner should work
on their pronunciation for this part of the sentence. The red is brighter the 
longer the length of the gap. Thus the gap \emph{-ale-} in \emph{talentos} `talents'
appears brighter than the gap \emph{-r-} in \emph{mostra} `show'.

\section{Voice contribution}

We have also implemented a voice contribution system allowing users to generate
and publish their own Omnilingo root indices. As with any networked system,
collecting and preserving data from our users can be done only with their
consent. Managing that consent within the context of a decentralised filesystem
comes with its own special challenges, and we designed what we think is as good
of a privacy- and consent-respecting system as we can.

As opposed to most current systems for data collection via crowd sourcing, in
Omnilingo, contributors own their own data and can define their own terms and
conditions for its use.

\subsection{Omnilingo privacy structures}

Our contribution privacy initiative brings with it a handful of new structures.
These are introduced bottom-up; read this section backwards if you prefer a
top-down introduction.

\subsubsection{Omnilingo session keys}

An Omnilingo session key is a JSON Web
Key\footnote{\url{https://datatracker.ietf.org/doc/html/rfc7517}}; our
implementation uses the SubtleCrypto
WebAPI\footnote{\url{https://developer.mozilla.org/en-US/docs/Web/API/SubtleCrypto}}
to generate and encode these keys. Currently we recommend only 256-bit AES-GCM
keys, and our Web client supports only this configuration.

Omnilingo session keys form the unit of "consent": for a given session key,
users may have contributed several samples. If a user wishes to revoke their
consent for a sample, they signal this by unpublishing the session key, thus
revoking consent for all samples contributed with that key.

For a more positive user experience, we recommend the user-facing interface
reference session keys by the pgpfone
wordlist\footnote{\url{https://web.archive.org/web/20100326141145/http://web.mit.edu/network/pgpfone/manual/index.html\#PGP000062}}
encoding of their fingerprint.

\subsubsection{Omnilingo encrypted object}

An Omnilingo encrypted object is an object which has been encrypted by an
Omnilingo session key; the structure is:

\begin{figure}
\begin{lstlisting}[language=json]
{ "alg": alg         // AesKeyGenParams
, "keyfpr": keyfpr   // key fingerprint: hexadecimal string encoding of the SHA-1 digest of the key
, "iv": iv           // initialisation vector used
, "encdata": encdata // Uint8Array of the encrypted data
}

\end{lstlisting}
\caption{An OmniLingo session key structure.}
\label{fig:session-key}
\end{figure}

We wrap in encrypted objects the MP3 of the contribution as well as the list of
Omnilingo clip structures.

Encrypted clip:
\begin{figure}
\begin{lstlisting}[language=json]
{ "chars_sec": chars_sec
, "clip_cid": CID(encrypt(clip_mp3))
, "length": length
, "meta_cid": meta_cid
, "sentence_cid": sentence_cid
}
\end{lstlisting}
\caption{An OmniLingo language index with encrypted clip MP3.}
\label{fig:clip-enc}
\end{figure}

\subsubsection{Omnilingo root with encrypted language index}

An Omnilingo root with encrypted language index is similar to the classic
Omnilingo root index: a JSON dictionary with language codes as keys, but
with Omnilingo encrypted language indices as the values.

An example:
\begin{figure}
\begin{lstlisting}[language=json]
{ "ab": { "cids": CID(encrypt(clip_list)) } }
\end{lstlisting}
\caption{An OmniLingo root index with encrypted language index.}
\label{fig:lang-enc}
\end{figure}

\subsubsection{Omnilingo encrypted root}

An Omnilingo encrypted root is a JSON dictionary; the keys are fingerprints of
Omnilingo session keys, and each value is the CID of an Omnilingo root with
language indices encrypted with the corresponding session key.

\begin{figure}
\begin{lstlisting}[language=json]
{ "ea6b0c9b..": "QmdzHipTQW.." }
\end{lstlisting}
\caption{An Omnilingo encrypted root. In this example, no decryption keys are provided.}
\label{fig:enc-root-nokey}
\end{figure}

Encrypted roots can optionally contain some of the referenced session keys, allowing
decryption. In Figure~\ref{fig:enc-root-onekey}, the key \texttt{ea6b0c9b...} is included.

\begin{figure}
\begin{lstlisting}[language=json]
{ "keys": {
    "ea6b0c9b..": JWK(key)
  }
, "dab24db6..": "QmWug9ie3b.."
, "ea6b0c9b..": "QmdzHipTQW.."
}
\end{lstlisting}
\caption{An Omnilingo encrypted root with components encrypted with different
keys. One key is provided for decryption, while the other is unavailable.}
\label{fig:enc-root-onekey}
\end{figure}

\subsubsection{Omnilingo identity}

An Omnilingo identity is a IPNS key (colloquially referred to as a \texttt{k5}).
Published to this \texttt{k5} is an encrypted root, containing the session keys for
which the user (the one controlling the private part of the \texttt{k5}). The
Omnilingo client has been updated to accept Omnilingo identities, fetching and
decrypting the contained encrypted indices.

In Figure~\ref{fig:enc-root-onekey} the material encrypted by session key
\texttt{ea6b0c9b2} can be used with the controlling user's consent, whereas the
material encrypted by session key \texttt{dab24db6} cannot be any longer, as the user
has unpublished the key.

\subsection{Data flows}

There are two new data flows introduced with this system: contributing data,
and retrieving contributed data.

\subsubsection{Contribution}

A contributor client will be drawing sentences from a (presumably classic)
Omnilingo language index, and contributing new clips. They start by generating
an Omnilingo identity (\texttt{k5}) and a session key. The session key is stored
locally.

When the user makes their first contribution (an MP3 recording of them reading
a sentence), a new Omnilingo encrypted root index is published to their \texttt{k5}:

\begin{figure}
\begin{lstlisting}[language=json]
{ "keys": {
    fpr(key): JWK(key)
  }
, fpr(key): CID({ // encrypted language index
    "XX": {
      "cids": [CID(encrypt([ // encrypted clip list
        encrypted_clip
      ]))]
    }
  })
}
\end{lstlisting}
\caption{An OmniLingo language metadata structure. Currently language
metadata consists of display names and reverse-keymaps.}
\label{fig:langmeta}
\end{figure}

As the user makes more contributions, the encrypted clip list grows in length,
updating the encrypted language index and encrypted root index, each time
republished to the \texttt{k5}, all under the same session key:

\begin{figure}
\begin{lstlisting}[language=json]
{ "keys": {
    fpr(key): JWK(key)
  }
, fpr(key): CID({ "XX": { "cids": [CID(encrypt(clip_list))] } })
}
\end{lstlisting}
\caption{An OmniLingo language metadata structure. Currently language
metadata consists of display names and reverse-keymaps.}
\label{fig:langmeta}
\end{figure}

At some point, the user decides to "roll" their session key, creating a new
session. (A client might decide to do this automatically, e.g. each time it is
opened, or each time the language is switched.) A new session key is generated,
and everything propagates up to the user identity (\texttt{k5}):

\begin{figure}
\begin{lstlisting}[language=json]
{ "keys": {
    fpr(key1): JWK(key1)
  , fpr(key2): JWK(key2)
  }
, fpr(key1): CID({ "XX": { "cids": [CID(encrypt(clip_list1))] } })
, fpr(key2): CID({ "XX": { "cids": [CID(encrypt(clip_list2))] } })
}
\end{lstlisting}
\caption{An OmniLingo language metadata structure. Currently language
metadata consists of display names and reverse-keymaps.}
\label{fig:langmeta}
\end{figure}

At some later time, the user decides to revoke consent to use the material
recorded under \texttt{key1}; the JSON Web Key encoded copy of \texttt{key1} is removed, only
\texttt{fpr(key1)} remains published under their identity:

\begin{figure}
\begin{lstlisting}[language=json]
{ "keys": {
    fpr(key2): JWK(key2)
  }
, fpr(key1): CID({ "XX": { "cids": [CID(encrypt(clip_list1))] } }) // consent revoked
, fpr(key2): CID({ "XX": { "cids": [CID(encrypt(clip_list2))] } })
}
\end{lstlisting}
\caption{An OmniLingo language metadata structure. Currently language
metadata consists of display names and reverse-keymaps.}
\label{fig:langmeta}
\end{figure}

Consumers who have stored \texttt{key1} will retain access to this data, just as they
would if they had stored the decrypted copies; however, use of it would
constitute a violation of the user's consent.

\subsubsection{Consumption}

Omnilingo consumers now have two types of root indices to deal with: classic
root indices and encrypted root indices. An encrypted root index may be
detected by the presence of the \texttt{keys} field; iterating over this dictionary
then gives the consumer a list of fingerprints to look up in the encrypted root
index, as well as the key needed to decode the resulting encrypted language
index.

\section{Future work}\label{sec:future}

We have presented our prototype web client and a proof-of-concept
terminal-based Python client. We would be happy to see native implementations
for major device platforms: graphical desktop operating systems as well as
smart phones and tablets. We leave as an open questions what sorts of
extensions can be made to smart watches, speakers, and VR systems.


There is a wide field of possible exercises that could be generated
based on OmniLingo data. We imagine that other sources of data could
be integrated, e.g. picture-word association from Wikidata (possibly
combined with Wiktionary), to build more comprehensive
language-learning applications. Distributing these over the IPFS
network should be straightforward.





Current client software stores user progress data locally, but modern
proprietary language-learning ecosystems allow multiple client
programs to sync, storing (or at least transferring) user progress
data through providers servers. We would like to implement a similar
user experience while respecting our users' privacy and fitting
organically into OmniLingo's philosophy of decentralisation. We are in
the design phase of implementing such a system.


Multiple OmniLingo root indices can be merged to allow for multiple
sources of language data to be used. Using the IPFS publish-subscribe
notification protocol, OmniLingo root indices could be advertised and
automatically collected and merged by clients. Designing a protocol
for such advertising (including how to resolve the omnipresent
question of trust, as well as domain-specific questions) remains a
future project.






\bibliographystyle{acl_natbib}
\bibliography{omnilingo-nlp4call}

\end{document}